\documentclass[letterpaper, 10 pt, conference]{ieeeconf}  

\IEEEoverridecommandlockouts                              

\usepackage{graphics} 
\usepackage{epsfig} 
\usepackage{times} 
\usepackage{amsmath} 
\usepackage{amssymb}  
\usepackage{hyperref}
\usepackage{bm}
\usepackage{cite}
\usepackage{multirow}
\usepackage{wrapfig}
\usepackage[caption=false, font=footnotesize]{subfig}
\usepackage{booktabs}
\usepackage{xcolor}
\title{\LARGE \bf
Motion Planning Transformers: A Motion Planning Framework for Mobile Robots
}

\author{Jacob J. Johnson $^{1}$, Uday S. Kalra$^{2}$, Ankit Bhatia$^{1}$, Linjun Li$^{1}{}^*$, Ahmed H. Qureshi$^{3}$, and Michael C. Yip$^{1}$ 
\thanks{$^{1}$Jacob J.J., Ankit B., and Michael C.Y. are with the Electrical and Computer Eng. Dept. at University of California San Diego,
{\tt\small \{jjj025, abhatia, yip\}@eng.ucsd.edu}}
\thanks{$^*$ Linjun L. was with the Electrical and Computer Eng. Dept. at University of California San Diego during this work.}
\thanks{$^{2}$Uday S.K. is with the Computer Science and Engineering Dept. at University of California San Diego,%
{\tt\small uday@ucsd.edu}}
\thanks{$^{3}$ Ahmed H.Q. is with the Computer Science Dept. at Purdue University,
{\tt\small ahqureshi@purdue.edu}}
}

\begin{document}

\maketitle
\thispagestyle{empty}
\pagestyle{empty}

\begin{abstract}
Fast and efficient sampling-based motion planning (SMP) is an integral component of many robotic systems, such as autonomous cars. A popular technique to improve the efficiency of these planners is to restrict search space in the planning domain. Existing algorithms define parametric functions to bound the search space, but these do not extend to non-holonomic robotic systems. Recent learning-based methods use a combination of convolutional and fully connected networks to encode the planning space. However, these methods are restricted to fixed map sizes, which are often not realistic in the real world. In this paper, we introduce a transformer-based approach, Motion Planning Transformer, to restrict the search space by learning to discern regions with a valid path from prior data. The model learns not only to restrict search spaces for simple 2D systems but also for non-holonomic robotic systems. We validate our method on various randomly generated environments with different map sizes and plan trajectories for a physical non-holonomic robot. We also provide a ROS2 plugin of our method for the Nav2 planning stack. The results show that our method reduces search space nodes by 2-12 times compared to traditional planners and has better generalizability than recent learning-based planners. 
\end{abstract}
\section{INTRODUCTION}
Sampling-based motion planning aims to construct a path by generating trees or graphs by randomly sampling points from the start to the goal location. In order to make the search more efficient, previous methods have restricted the search space for generating the shortest paths \cite{6942976}, but these methods are based on parametric functions and would be infeasible for robots with kinematics \cite{li2016asymptotically} or safety constraints \cite{5970128, 6224727}. Recently, learning-based planning methods have tried to bridge these gaps by learning to sample points satisfying kinematic constraints for guiding the search \cite{li2021mpc, 9341283}. However, learning-based methods often combine convolution neural networks (CNN) and multi-layer perceptrons (MLP) to guide the search tree \cite{qureshi2020motion, 9341283, kumar2019lego}, making them restrictive to fixed environment sizes. On the other hand, methods such as Value Iteration Network (VIN) \cite{tamar2016value}, which extends to different environment sizes, perform poorly in larger environments \cite{nardelli2018value}. Hence, in our ever-expanding pursuit of automation, we need techniques that can scale up while effectively keeping the planning problem’s complexity and computation time in check. We show that transformer-based models can meet these requirements while overcoming the challenges of prior learning-based methods.

\begin{figure}[t]
    \centering
    \includegraphics[width=\linewidth]{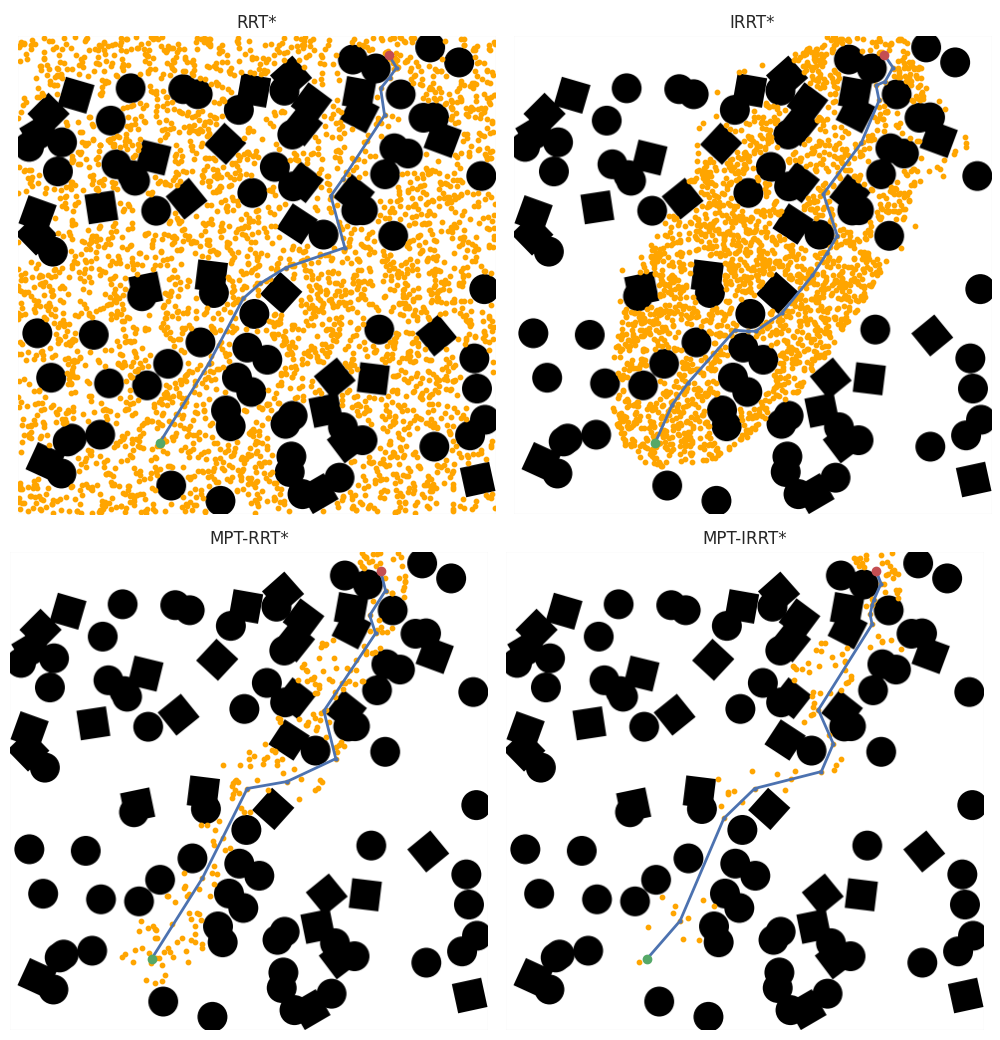}
    \vspace*{-2em}
    \caption{(From top left clockwise) Vertices used by RRT*, IRRT*, MPT-IRRT*, and MPT-RRT* for the same start (green) and goal (red) positions. MPT aided planners are able to reduce significantly the number of vertices (orange) required to search for a path.}
    \label{fig:mpt_sampling}
    \vspace{-1em}
\end{figure}
\begin{figure}[!t]
    \centering
    \includegraphics[width=\linewidth]{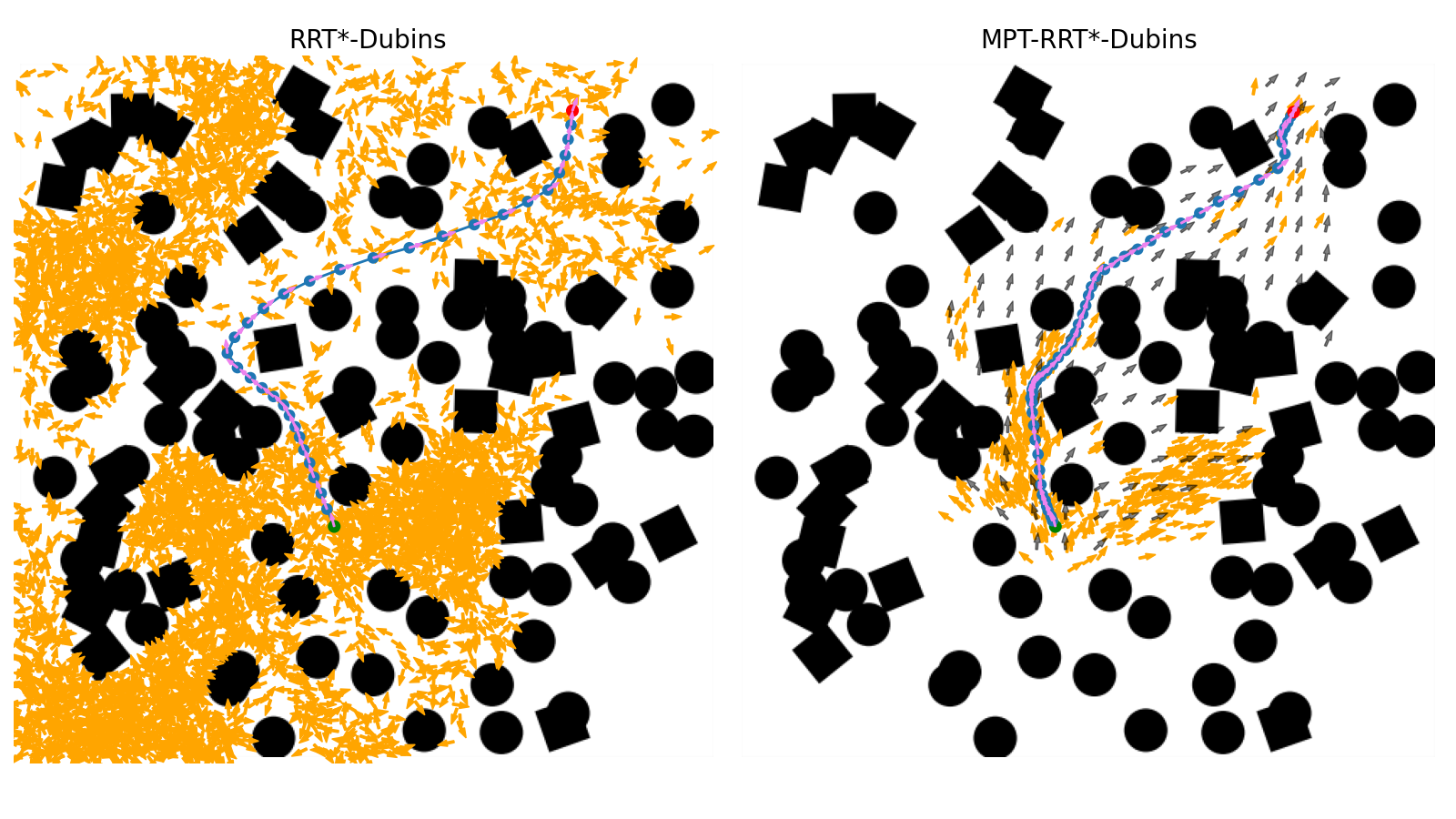}
    \vspace*{-2em}
    \caption{Vertices used by RRT* (left) and MPT-RRT* (right) for the same start (green) and goal (red) positions in the $SE(2)$ space. By guiding the sampling around anchor points (gray), the planner is able to achieve a shorter path with far fewer samples.}
    \label{fig:car_samp}
    \vspace{-2em}
\end{figure}

Motion planning and language translation tasks require a good understanding of global dependencies. The syntactic and semantic structure of a sentence is only inferred by reading the entire sentence. Similarly, the trajectory of a local plan is influenced by the orientation of far-away obstacles. The ability of transformers to learn these long horizon dependencies has made them the powerhouse of natural language processing \cite{nips_attention}. Hence transformer models seem ideal for making long horizon correlations for path planning problems.

In this paper, we introduce Motion Planning Transformer (MPT), a transformer-based model to reduce search space for planning algorithms. The following are our main contributions 
\begin{itemize}
    \item A framework for using Transformers for planning for mobile robots. We assess the performance of MPT aided SMPs \cite{karaman2011sampling,6942976} on synthetic datasets representing cluttered and long-horizon environments. Our results illustrate that MPT assisted planners achieve a 7-28\% improvement in accuracy over recent learning-based planners while matching the accuracy of traditional planners. MPT planners also reduce vertices on the planning tree by 2-12 times and planning time by 7-25 times compared to traditional planners (See Fig. \ref{fig:mpt_sampling}).
    \item Introduce a novel training routine, allowing generalization to larger maps. This training routine improves the planners accuracy by 60\% for larger maps.
    \item Expand the framework for planning for $SE(2)$ robots. We adapt our model to predict orientations enabling us to reduce the search space for $SE(2)$ robots. MPT aided planners are able to reduce planning time by 2 times compared to traditional planners (See Fig.  \ref{fig:car_samp}).
    \item Provide a ROS2 plugin for the Nav2 navigation stack \cite{9341207} for our method \footnote{The package will be made available after the review process}. This will be beneficial for the robotics community to work with and extend our models.
\end{itemize}

\section{Related Works}
The most relevant work to our transformers-based region proposal network for motion planning is perhaps the guided sampling-based motion planning methods. They analytically or through learned heuristics determine a subset in robot space that probably contains a path solution. For instance, \cite{qureshi2016potential, tahir2018potentially} employ Artificial Potential Fields (APF) within sampling-based methods such as RRT* \cite{karaman2011sampling} and Bidirectional RRT* \cite{qureshi2015intelligent} to guide a subset of random samples towards promising regions that possibly contain an optimal path solution. In contrast, Informed-RRT* (IRRT*) \cite{6942976} depends on an initial path from an RRT* algorithm to compute an ellipsoidal region probably containing an optimal path solution. However, in most planning problems finding an initial path solution is itself challenging. In a similar vein, Batch Informed Trees (BIT*) \cite{gammell2015batch} begins from an elliptical region formed by a straight line path ignoring all obstacles and incrementally expand it until an initial path solution is found. Once an initial path is determined, it is further optimized by adapting the precomputed ellipsoid and generating new samples within that space. 

Many have also used learning-based methods to reduce search spaces. 
\cite{8412538, 9561104} uses Gaussian Mixture Models (GMM) to learn a distribution from prior motion plans. \cite{8412538} constructs a roadmap from previous solution paths, and uses the learned roadmap to plan in relatively similar environments to the training data. \cite{9561104} learns a set of local samplers, parameterised as GMM, from prior datasets and decomposes a new planning scene into these local samplers. A global plan is constructed by sampling from these learned local primitives to construct the path. For cluttered spaces these planners would need to need to use GMM's with a high mode value while for large planning spaces the decomposition would be non-trivial. \cite{doi:10.1177/0278364918781001,DBLP:conf/iclr/ChenDLYLS20} used a learned value function to guide the graph search, while \cite{kumar2019lego} learned a generative model to sample points along bottleneck regions. Similarly, Value Iteration Networks (VIN) \cite{tamar2016value} discretizes the space and learns a value map to guide path planning. Universal Planning Networks (UPN) \cite{srinivas2018universal} extends VIN to continuous control spaces. 
These methods are often difficult to train and interpret, and many of them are yet to be evaluated in real world navigation tasks.


Neural Motion Planning \cite{qureshi2020motion,ichter2018learning} has recently emerged as a promising tool for solving a wide range of planning problems under various task constraints, ranging from non-holonomic \cite{9341283,li2021mpc} to advanced manifold kinematic constraints \cite{qureshi2020compnet}, with high computational speed. These methods learn sampling distributions from expert demonstrations and, on execution, generate samples for an underlying planner forming a subset that potentially contains a path solution. However, these approaches assume a fixed size input environment map and often require redefining network architectures and retraining for different map sizes. Recent developments in deep learning, primarily through Transformers \cite{vit, liu2021swin},  have provided us with ways to relax such assumptions. Our proposed approach leverages these developments and introduces a region proposal framework that can work with variable map sizes and enhances underlying motion planners to solve complex problems in cluttered environments. 

\section{Method}
\begin{figure*}[t]
    \vspace{0.5em}
     \centering
     \includegraphics[width=\textwidth]{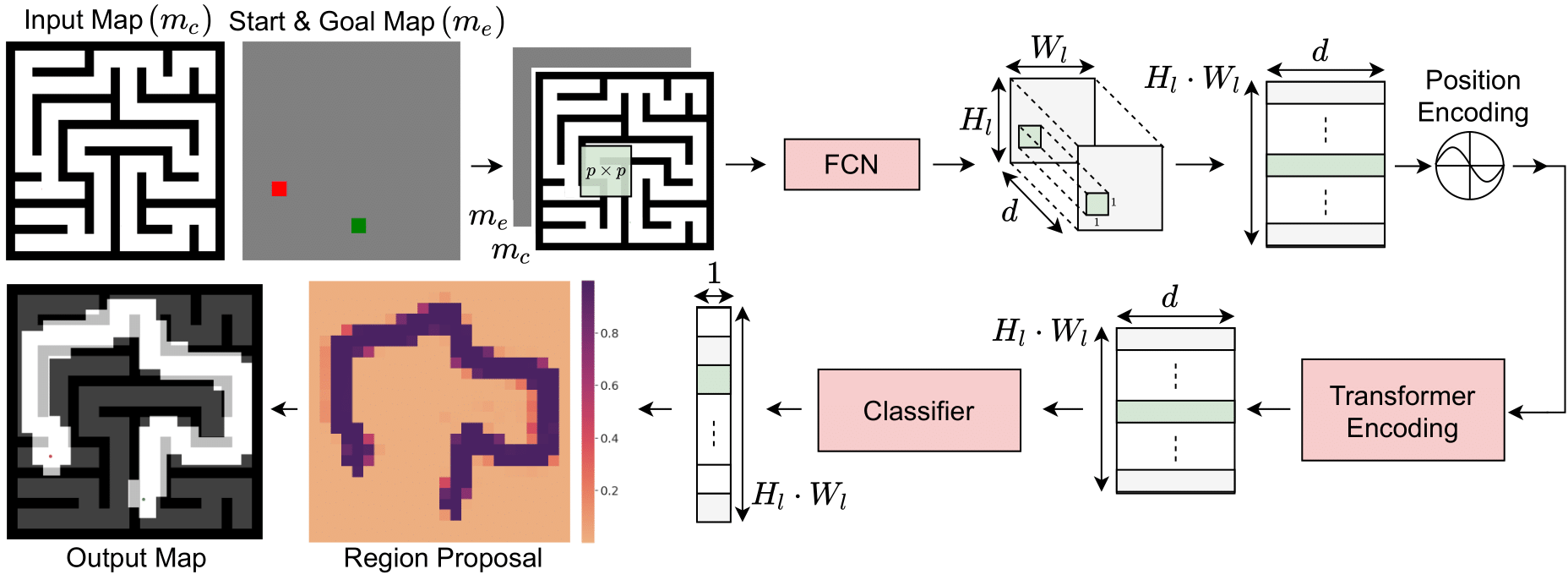}
     \vspace*{-1.5em}
     \caption{Overview of MPT Module for planning in $\mathbb{R}^2$. (Start from the top left and move clockwise) The input map, and the start (green) and goal (red) encoded map are concatenated together and passed as inputs to the model. The Fully Convolution Network (FCN) passes a sliding window of size $p\times p$ over the input and encodes each patch into latent vectors of dimension $d$. After reshaping, fixed positional encodings are added to the latent vectors to inject spatial location. The modified latent representation is used by the transformer module to identify patches through which a path might exist. This information is encoded into a latent vector of size $d$, which is used by a classifier to provide the probability that a path might pass through the patch. The patches with probability greater than 0.5 are used to create the mask for the map. The light green shading highlights the flow of information for a single patch from the input to the output of the model.}
     \label{fig:NetArchitecture}
     \vspace{-2em}
\end{figure*}

Given a start ($x_s$) and goal ($x_g$) state, the objective is to propose a sequence of states $x_i\in \mathcal{X}$ for $i\in\{0, 1, \ldots n\}$ such that $x_0=x_s$, $x_n\in \mathcal{X}_{\text{goal}}$ and the trajectories joining these states should be both kinematically feasible and collision free. Here, $\mathcal{X}$ represents the robot state space and $\mathcal{X}_g \triangleq \{x: \|x-x_g \|\leq \epsilon, x\in \mathcal{X}\}$ for a user defined threshold ($\epsilon$). The set of motion planning problems we will focus on this paper is 2D navigation hence $\mathcal{X}\in \{\mathbb{R}^2, SE(2)\}$. MPT proposes promising regions in $\mathcal{X}$ where the underlying planners can search for path solutions. In the following sections we describe our method in detail.

\subsection{Motion Planning Transformer (MPT)}

The MPT module is a region proposal network that uses a transformer network to identify regions of interest. An overview of the model is shown in Fig. \ref{fig:NetArchitecture}. The input to the model is a representation of the planning scene, and an encoding of the start and goal points. The planning scene is represented using an occupancy matrix, $\bm{m_c}\in \mathbb{R}^{H\times W}$, where an element with 1 indicates an occupied space and 0 denotes free space. In some navigation problems, these representations are also called costmap and have additional cost terms associated with safety and robot constraints. The start-goal encoding of the planning problem is formed by highlighting patches of size $p\times p$ on a tensor of size $H\times W$ with values -1 and 1 for the start and goal point respectively. For $SE(2)$ space, we add two additional matrices to represent the robot orientation which is represented using sine and cosine values respectively. These matrices are concatenated to form a tensor $\bm{m}$ and passed to the feature extractor.

\textbf{Feature Extractor:} 
The feature extractor is a Fully Convolution Network (FCN) that encodes the environment and the given planning problem into a latent space. As shown in prior works \cite{vit, NIPS2015_14bfa6bb}, the feature extractor reduces the dimensionality of the input space by using a series of convolution, ReLU, and MaxPool layers. The FCN passes a sliding window of size $p\times p \times i$ over $\bm{m}$ to generate an output of size $H_l \times W_l \times d$, where $H_l$ and $W_l$ is determined by the size of the costmap and the FCN, and $d$ is the latent dimension of the transformer encoder. $i$ is 2 for the $\mathbb{R}^2$ space and 4 for the $SE(2)$ space. The patch size $p$ defines the size of local structures that the model encodes into the latent vector and also the smallest area the model can mask. Larger patch sizes will have lower resolution of masking thus the resulting planner would perform similar to traditional planners.

The output of the FCN is reshaped row-wise to size $(H_l\cdot W_l) \times d$ and fed to the position encoder. Hence each latent vector at index $(i_F, j_F)$ is mapped to the row $i_{F}*W_l+j_{F}$ where $i_F\in\{0, 1, \ldots, H_l-1\}$ and $j_F \in \{0, 1, \ldots, W_l-1\} $. Each row vector of this matrix corresponds to a $p\times p$ patch that the sliding window moves over and is referenced by an anchor point similar to Faster R-CNN \cite{NIPS2015_14bfa6bb}. We choose the 2D co-ordinate corresponding to the center pixel of the patch as the anchor point.

\textbf{Position Encoding:}
Transformer and convolutional models are agnostic to the spatial location of their inputs \cite{vit}. A common solution is to add learned or fixed vectors to encode the position of each input \cite{10.5555/3305381.3305510}. We used fixed vectors to encode the position and used the following for testing,
\begin{align}
    PE(j, 2i)   &= \sin\left(\frac{j}{10000^{2i/d}}\right) \label{eqn:pe_even}\\
    PE(j, 2i+1) &= \cos\left(\frac{j}{10000^{(2i+1)/d}}\right) \label{equn:pe_odd}
\end{align}
$j\in\{0, 1,\ldots, H_l-1, \ldots i_F*W_l+j_F, \ldots,  H_l*W_l-1\}$ , and $i\in\{0, 1, 2 \ldots, d-1\}$ similar to \cite{nips_attention}. The maximum value $j$ could take was set at $H_{max}^2-1$. For maps larger than the training data, we observed that models trained using the position encoding in Eqn. \ref{eqn:pe_even} and \ref{equn:pe_odd} created a bias that prevented MPT from selecting regions of the state space outside the training map size. One way to resolve this overfitting is to train on maps of various sizes to ensure that the network observes different position encoding. However, the data collection is often computationally expensive, especially for larger maps. Instead, to overcome this bias, we leverage the fact that a proposed plan is invariant to linear translation of the state space, and train our model by randomly shifting the position encoder. Hence for training we randomly sample $(i_R, j_R)\sim U[0, H_{max}-1]^2$ where $U[0, H_{max}-1]$ is a uniform distribution over discrete set of integers from $0$ to $H_{max}-1$ and use the following position encoder for each map:
\begin{align*}
    PE(j, 2i)   &= \sin\left(\frac{j+k}{10000^{2i/d}}\right) \\
    PE(j, 2i+1) &= \cos\left(\frac{j+k}{10000^{(2i+1)/d}}\right)
\end{align*}
where if $j=i_F*W_l+j_F$ then $k=H_{max}*(i_R+i_F)+j_R$. This effectively translates the map to different position encoding values during training and ensures that the model is not biased by the size of the training map. The modified latent vector is then passed to the transformer encoder.

\textbf{Transformer Encoder:}
The transformer module is responsible for learning the connections between the different local regions on the map for a given planning problem. It infers these connections by passing the latent vectors through a series of multi-headed self-attention (MSA) and multi-layer perceptron (MLP) blocks. Between every MSA and MLP block we apply Dropout, Layer Norm and residual connections similar to \cite{nips_attention} and \cite{vit}. We also add gradient checkpoints \cite{chen2016training} after the MSA blocks to be more memory efficient. Hence by attending to all patches, the network encodes the importance of each patch to the given planning problem in its output.


\textbf{Classifier:}
Using the encoded importance, the classifier predicts if each of the anchor point is of interest to the current planning problem. This is efficiently implemented using a $1\times 1$ convolution layer, similar to Faster R-CNN \cite{NIPS2015_14bfa6bb}. A mask is generated by setting $p\times p$ patch, centered around selected anchor points, to 1 on a matrix of size $H\times W$.

\textbf{Orientation Prediction:}
For the $SE(2)$ space, each positive anchor point is assigned an orientation which is generated by this layer. Like the classifier, this is also implemented efficiently using a $1\times1$ convolution layer.

\subsection{Path Planning}
Any traditional or learning-based planner can be used to find the path by searching the masked region. For the point robot we sample uniformly across a square grid centered around each anchor point, while for the Dubins car we sample uniformly across an ellipse whose major axis is oriented along the predicted orientation. In this work, we use variations of the Rapidly Exploring Random Trees (RRT) algorithm, that guarantee optimality \cite{karaman2011sampling,doi:10.1177/0278364915614386,6942976}, to find the path using the sampled nodes. 

In rare occasions, due to mis-classifications, MPT guided planners fail to generate a solution. To overcome these errors, we alternate between searching the masked (exploitation) and unmasked (exploration) regions of the map. We show that this technique is able to preserve the benefits of MPT guided planners while achieving higher efficiency in planning.

\section{Experiments}
We evaluated the planning capabilities of MPT aided SMPs and compared them with both traditional and learning-based planners. In the following section, we will go over our environment and robot setup, model training, and results.

\begin{figure*}[t]
    \centering
    \includegraphics[width=\textwidth]{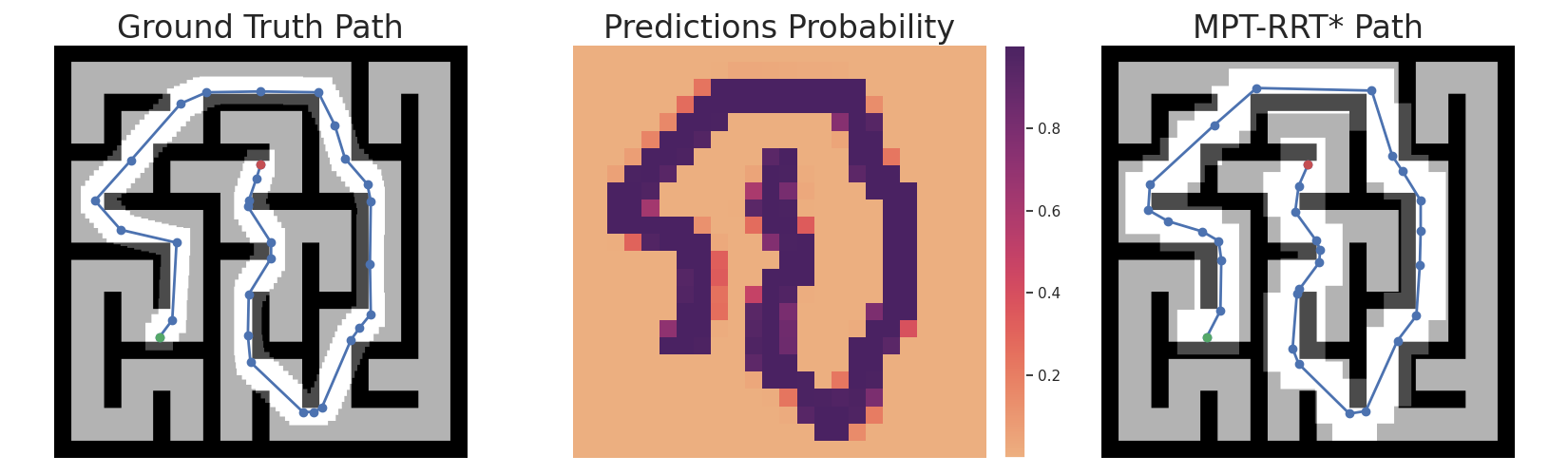}
    \vspace*{-1.5em}
    \caption{Planned path for the Maze environment. Left: Ground truth path for a given start (green) and goal (red) point from the validation data. Center: The prediction probability from the MPT. Right: Masked map and planned trajectory for the given start (green) and goal (red) point using MPT-RRT*.}
    \label{fig:pointRobotPlannedPath}
    \vspace{-2em}
\end{figure*}

\subsection{Setup}
\textbf{Environments:}
To test the planning capabilities of our method, we evaluated the model on randomly generated maps from two different classes of environments. The first environment is called the Random Forest, where circular and square objects are randomly placed on the map. It simulates real world scenes with narrow passages and crowded spaces. The second environment is called the Maze environment. Each map from this environment is a perfect maze, generated using randomized depth-first search. A characteristic of a perfect maze is that any start and goal pairs on this map are reachable by a collision-free path. The maze environment mimic long-horizon planning problems because even if the start and goal are geometrically close, the planner would have to take a circuitous path to reach the goal. For all environments we use an occupancy map of 5cm per pixel to check if the robot is in collision.
\label{sec:env}

\begin{figure*}
    \subfloat{\includegraphics[width=0.49\textwidth]{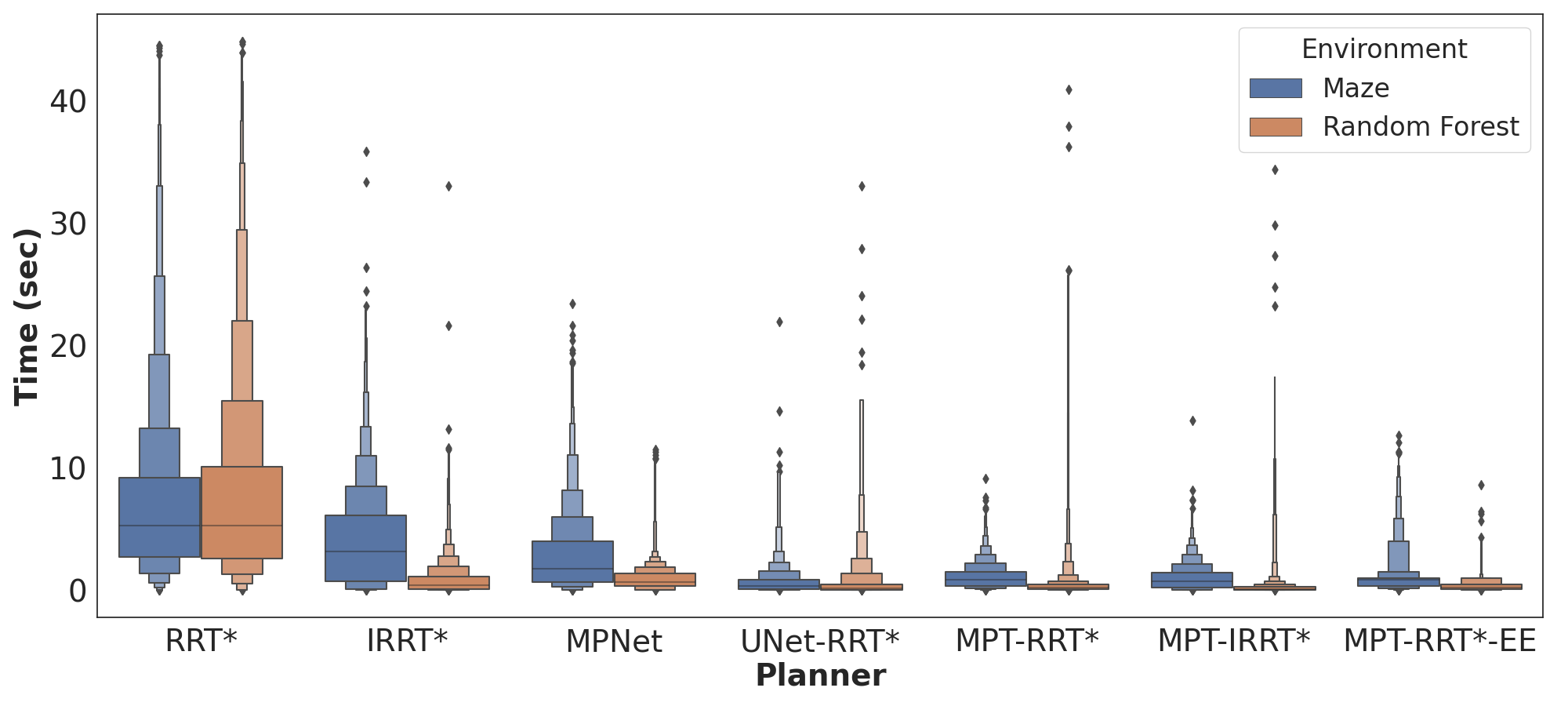}}
    \hfil
    \subfloat{\includegraphics[width=0.49\textwidth]{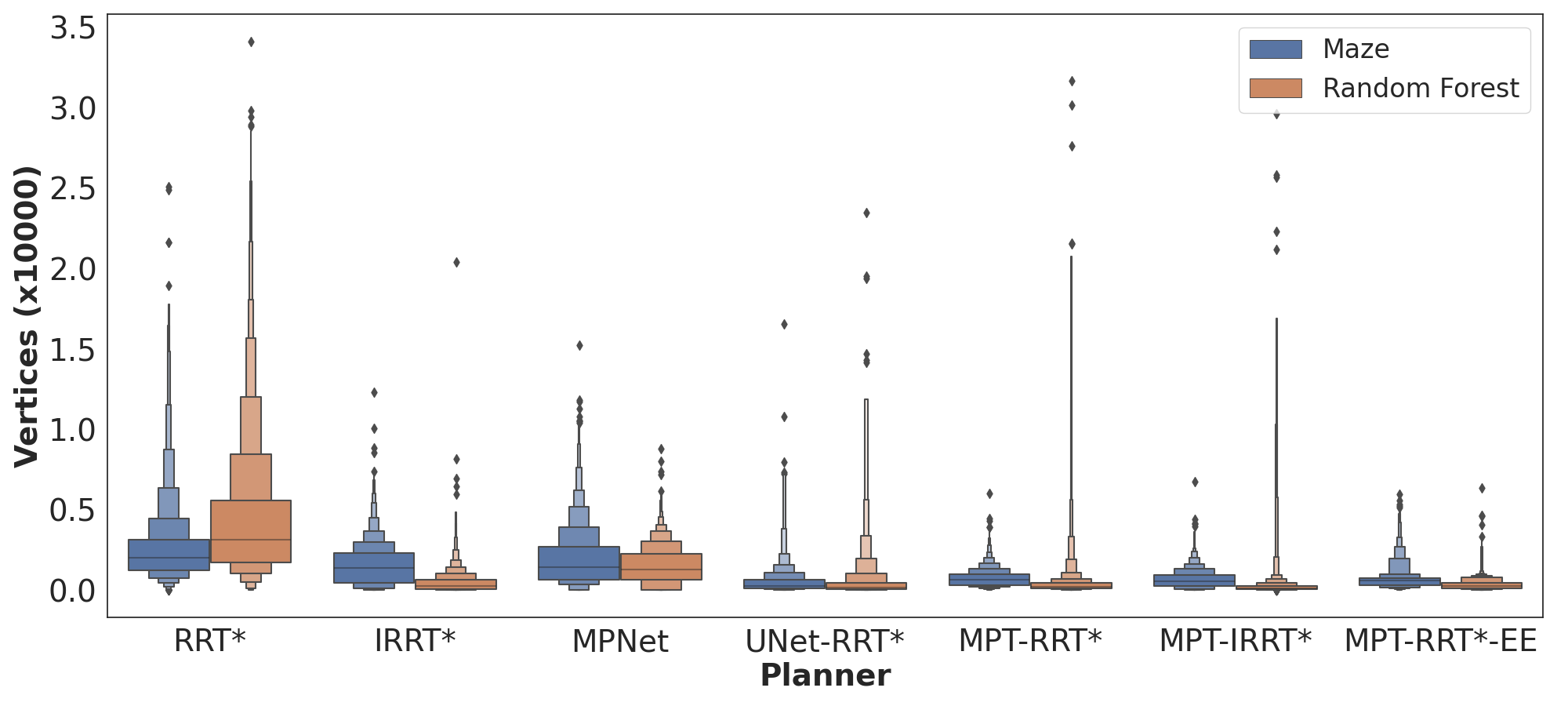}
    }
    \vspace*{-1em}
    \caption{Planning statistics for the Point Robot Model. Left: The planning time for traditional and learning-based planners. Right: Number of vertices in the planning tree for traditional and learning-based planners. MPT aided planners consistently reduce the planning time and the vertices in the planning tree, resulting in a lower variance of these statistics for these planners.}
    \label{fig:pointRobotStats}
    \vspace{-1.25em}
\end{figure*}
\textbf{Robot Models}
We test our algorithm on two kinds of robotic systems that encapsulate a large set of mobile systems. The first is a simple Point Robot Model that can move in any direction in its state-space $\mathcal{X}=\mathbb{R}^2$. Solutions for this model can be easily extended to a large number of indoor and outdoor robots. For such robots, taking the Minkowski sum of the obstacles and the robot’s footprint reduces it to a point robot \cite{1676196}. 
The other robot we tested is a Dubins Car Model with state-space $\mathcal{X}=SE(2)$. 

\subsection{Training}
The MPT model is trained in an end-to-end fashion under supervision similar to \cite{girshick2015fast}. Each mini-batch is formed from a single planning problem containing positive and negative anchor points. We define positive anchor points as those within 0.7m distance to the trajectory, while all other anchor points are considered negative samples. For training, negative anchor points are randomly chosen to have a ratio of 1:1 of positive and negative samples. For the Dubins Car Model, each positive anchor point had an orientation associated with it. This orientation was the average of all the orientations of the trajectory within 0.7m of the path.

We trained the network by minimizing the cross-entropy loss for the anchor points and maximizing cosine similarity measure for the orientation using the Adam optimizer \cite{DBLP:journals/corr/KingmaB14} with $\beta_1 = 0.9$, $\beta_2=0.98$, and $\epsilon=1e^{-9}$. We varied the learning rate as proposed in \cite{nips_attention} with warm-up steps of 3200. Each model was trained for 100 epochs with a batch size of 128. The models were trained on one machine with 4 NVIDIA 2080GTX graphics card. The MPT model for the Point Robot took 21hrs and for the Dubins Car took 12hrs to train.

\textbf{Datasets:}
We trained an MPT model for the Point Robot on 1750 randomly generated maps from each environment mentioned in Section \ref{sec:env}. For each map, 25 paths were generated using the RRT* planner that terminated after 90 seconds. Similarly we trained an MPT model for the Dubins Car on 900 randomly generated maps from the Random Forest environment. We collected 25 paths from each map using the RRT* planner using Dubins curve \cite{Dubins_1957} as the node connector. All the maps used were of size $480\times 480$.

\subsection{Point Robot Model}
For the Point Robot Model, we compared MPT-aided planners with traditional and learning-based planners. We also looked at the capabilities of other image segmentation approaches such as UNet \cite{ronneberger2015unet} to highlight the area where a potential path might exist. We choose UNet because, like MPT, it can generalize to maps of different sizes. We call the aided planners Y-X, where Y is the underlying method used to generate the mask and X is the SMP planner. MPT-RRT*-EE represents the MPT aided RRT* planner with the exploration and exploitation strategy (hybrid planning).

\begin{table}
    \caption{Comparing planning accuracy, and median time and vertices for the Point Robot Model on unseen environments of the same size as the training data for Random Forest.}
    \label{tab:pointRobot_480_forest}
    \centering
    \begin{tabular}{lccc}
        \toprule
        Algorithm   & Accuracy & Time (sec) & Vertices \\
        \midrule 
        RRT*        & 100\%   & 5.448 & 3228 \\
        IRRT*       & 100\%   & 0.425 & 267  \\
        BIT*        & 100\%   & 0.477 & 819 \\
        UNet-RRT*   & 30.27\% & 0.167 & 168  \\
        UNet-RRT*-EE& 100\%   & 2.58  & 1913 \\ 
        MPNet       & 92.35\% & 0.296 & 63   \\
        MPT-RRT* (ours)    & 99.40\% & 0.194 & 233  \\
        MPT-IRRT* (ours)  & 99.40\% & 0.087 & 136  \\
        MPT-RRT*-EE (ours)  & 100\%   & 0.211 & 247  \\
        \bottomrule
    \end{tabular}
    \vspace{-2em}
\end{table}

\begin{table}
    \caption{Comparing planning accuracy, and median time and vertices for the Point Robot Model on unseen environments of the same size as the training data for Maze.}
    \label{tab:pointRobot_480_maze}
    \centering
    \begin{tabular}{lccc}
        \toprule
        Algorithm & Accuracy & Time (sec) & Vertices \\
        \midrule 
        RRT*              & 100\%   & 5.364 & 2042 \\
        IRRT*             & 100\%   & 3.139 & 1394 \\
        BIT*              & 100\%   & 2.870 & 2002 \\
        UNet-RRT*         & 21.4\%  & 0.346 & 277  \\
        UNet-RRT*-EE      & 100\%   & 4.133 & 2139 \\ 
        MPNet             & 71.76\% & 1.727 & 1409 \\
        NEXT-KS           & 28.27\% & 3.021 & 387 \\
        MPT-RRT* (ours)   & 99.16\% & 0.870 & 626 \\
        MPT-IRRT* (ours)  & 99.16\% & 0.784 & 566 \\
        MPT-RRT*-EE (ours)& 100\%   & 0.869 & 585 \\
        \bottomrule
    \end{tabular}
    \vspace{-2em}
\end{table}

The first set of experiments examined the network's ability to generalize to unseen maps of the same dimension as the training data. We compared the planners on 2500 random start and goal pairs for maps from the Maze and Random Forest environment. For each planner, we report the 1. Accuracy - the percentage of planning problems the planner solves,  2. Time (sec) - the amount of time it takes to generate the mask (if applicable) and plan a path shorter or of equal length to a path from the RRT* planner searching for 90 seconds, and 3. Vertices - the number of collision-free states sampled by the planner to construct the planning tree. The summary statistics of the experiment are reported in Table \ref{tab:pointRobot_480_forest} and \ref{tab:pointRobot_480_maze}. We see that MPT aided planners reduce the planning time and vertex count of the planning tree substantially. In Fig. \ref{fig:pointRobotPlannedPath}, we show an example of a planned path from the maze environment. 

\begin{figure*}[t]
    \begin{minipage}[b]{0.395\linewidth}
        \centering
        \includegraphics[width=\linewidth]{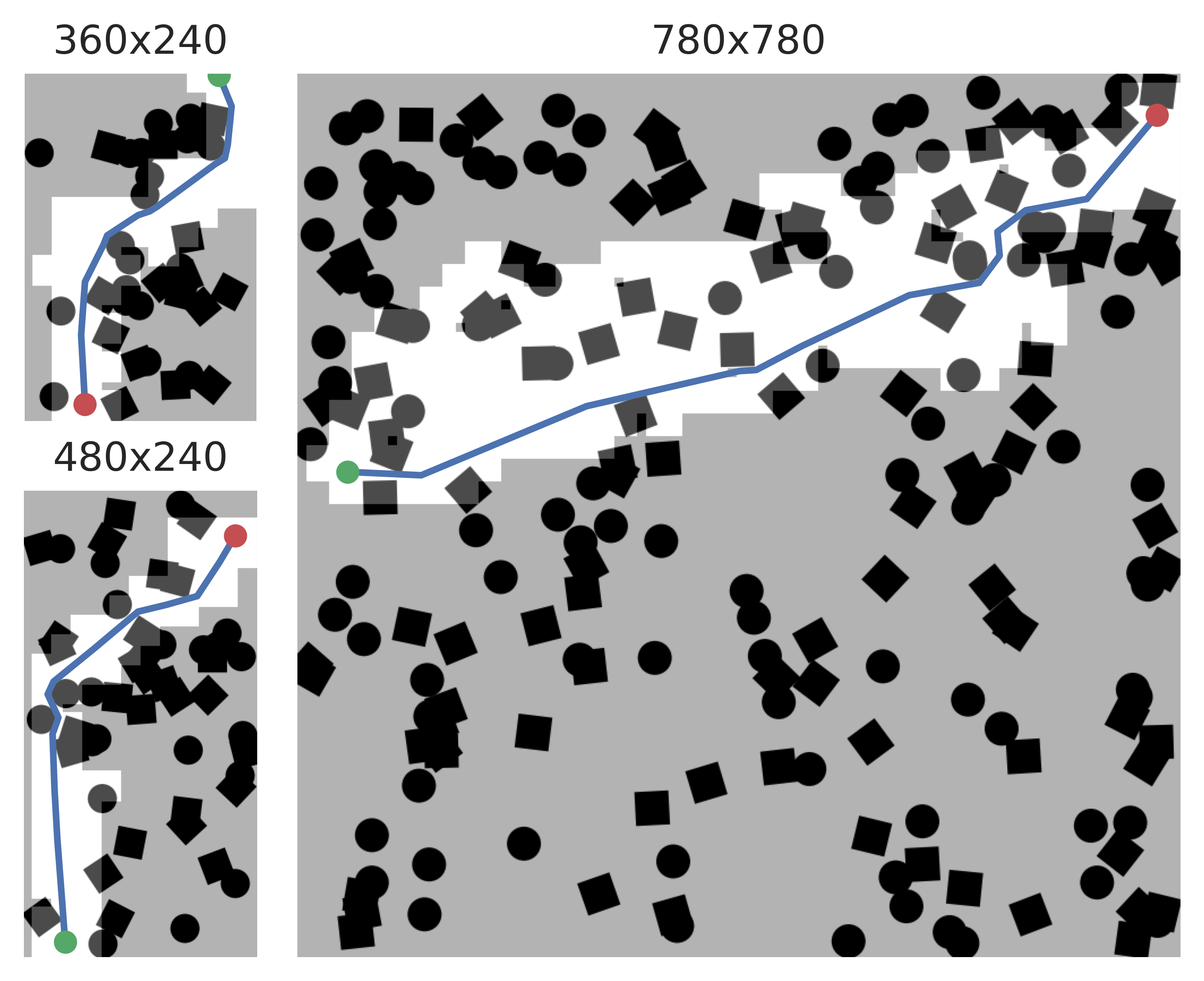}
        \vspace*{-1.5em}
        \caption{Plot of paths for Random Forest environments of different size. The architecture of the MPT Model allows flexibility in planning for environments of different sizes.}
        \label{fig:different_mapSize}
    \end{minipage}
    \hfil
    \begin{minipage}[b]{0.595\linewidth}
        \centering
        \subfloat{\includegraphics[width=0.5\linewidth]{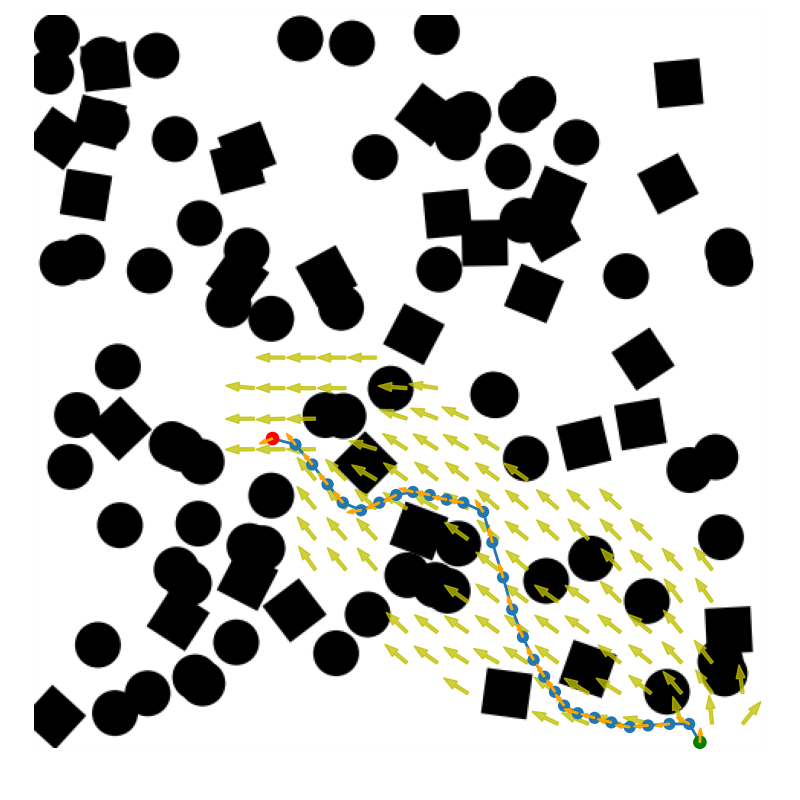}}
        \subfloat{\includegraphics[width=0.5\linewidth]{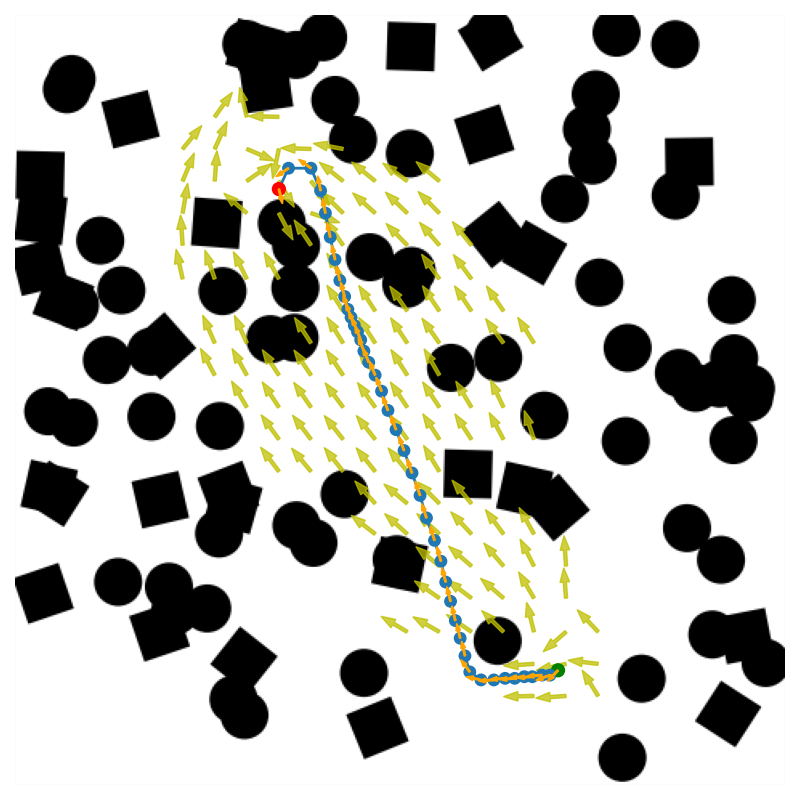}}
        \vspace*{-1.5em}
        \caption{MPT can also be trained to aid SMP planners for non-holonomic robots. Left and Right: Planned paths on Random Forest environment using MPT-RRT*. MPT identifies regions in $SE(2)$ through which a non-holonomic path exists.}
        \label{fig:dubins_car_map}
    \end{minipage}
    \vspace{-0.5em}
\end{figure*}

\begin{table*}[t]
    \caption{Comparing planning accuracy, and median time and vertices for Point Robot on maps of the different sizes.}
    \centering
    \begin{tabular}{cccccccccc}
        \toprule
        Map Size & & RRT* & IRRT* & BIT* & MPT-RRT* & MPT-IRRT* & MPT-RRT* & MPT-IRRT* & MPT-RRT*-EE\\
        (\# Obstacles) & & & & & (F.E.) & (F.E.) & & &  \\
        \midrule
        \multirow{2}{*}{360$\times$240}  & Accuracy   & 100\%  & 100\%  & 100\% & 97.4\%  & 97.4\% & 99.20\% &  99.20\% & 100\%\\
                                         & Time (sec) & 5.926  & 0.286  & 0.625 & 0.265   & 0.062 & 0.248   & 0.054    & 0.297 \\
                (35)                     & Vertices   & 3660   & 257    & 1069  & 377     & 118   & 354     & 106      & 382  \\
        \midrule
        \multirow{2}{*}{480$\times$240}  & Accuracy   &  100\%  &  100\% & 100\% & 96.3\%  & 96.3\% & 98.5\% & 98.5\% & 100\%\\
                                         & Time (sec) &  6.308  &  0.394 & 0.590 &  0.268  & 0.072  & 0.265  & 0.073  & 0.302 \\
                (50)                     & Vertices &  3480     &  291   & 1061  &  348    & 130 & 319    & 131    & 362  \\
        \midrule
        \multirow{2}{*}{560$\times$560}  & Accuracy   & 100\%  &  100\% & 100\% & 75.6\%  & 75.6\% & 99.7\% & 99.7\% & 100\%\\
                                         & Time (sec) &  6.725 &  0.283 & 0.397 & 0.253   & 0.082  & 0.181  & 0.083  & 0.217 \\
                (100)                    & Vertices   &  3854  &  203   & 810   &  262    & 101    & 218    & 112    & 237  \\
        \midrule
        \multirow{2}{*}{780$\times$780}  & Accuracy   & 100\% & 100\% & 100\% &  37.9\% & 37.9\% & 99.5\% & 99.5\% & 100\% \\
                                         & Time (sec) & 8.095 & 0.476 & 0.542 &  0.274  & 0.095 & 0.297  & 0.152  & 0.285\\
                (200)                    & Vertices   & 4292  & 255   & 974   &  284    & 142   & 241    & 161    & 238  \\
    \bottomrule
    \end{tabular}
    \label{tab:different_mapSize}
    \vspace{-2em}
\end{table*}

To better understand the advantages of MPT, we visualize the distribution of the planning time and vertices in Fig. \ref{fig:pointRobotStats} for the Random Forest and Maze environment using Letter-value plots. These plots help to observe the tail of the distribution of the metrics. Naive RRT* has a heavier tail distribution than MPT-RRT* because for start and goal pairs further away from each other, the planner needs to generate a denser graph to search a larger space, requiring more time and vertices.  On the other hand, MPT-aided planners focus their search near regions highlighted by the model, and as a result, they plan faster with fewer vertices. We see a thin tail distribution for MPT-RRT* and MPT-IRRT* planners for the random forest environment. This is because the planner only terminates when the cost of the path is below a certain threshold. Even if a solution is found, MPT planners continue to search to reduce the path length, resulting in longer planning times and vertex count for few trajectories. The number of such problems accounts for less than 0.75\% of the planning problems.

We also observe that IRRT* and BIT* achieves similar planning time and planning tree vertices compared to the aided planners. This is because these planners, like MPT, reduces the planning search space once an initial solution is found by bounding the initial path with an Ellipse. Such heuristics do not work for long-horizon problems like the Maze environment, and MPT aided planners outperform traditional methods.


The MPT planners also outperform other learning-based approaches. Planners that used UNet to propose patches performed poorly. We believe this is because the convolution layers can only learn the connections between local patches, and deeper networks would be required to learn global connections. As a result, it fails to highlight the area of interest for the given planning problem. We also tested the recently proposed Exploration-Exploitation Tree with kernel smoothing (NEXT-KS) \cite{DBLP:conf/iclr/ChenDLYLS20} on the Maze environment. The same validation set from Table \ref{tab:pointRobot_480_maze} was used, but the map size was reduced from 480$\times$480 pixels to 16$\times$16 pixels. The drop in performance can be attributed to the value function proposing samples that are locally optimum. We do not compare the planners performance on the forest environment because reducing the map size reduces the distance resolution from 0.05meter/pixel to 0.75meter/pixel, which if applied to the Forest environment would oversimplify collision free regions. MPNet, on the other hand, performs considerably better than UNet-RRT* for both environments, but not as well as MPT planners. We attribute the weak generalization to the lack of training data. In  \cite{qureshi2020motion}, the authors used nearly 400k trajectories to train their model, whereas we only provided around 88k trajectories for all our models.

\begin{figure*}[t]
    \vspace{1em}
    \includegraphics[width=\textwidth]{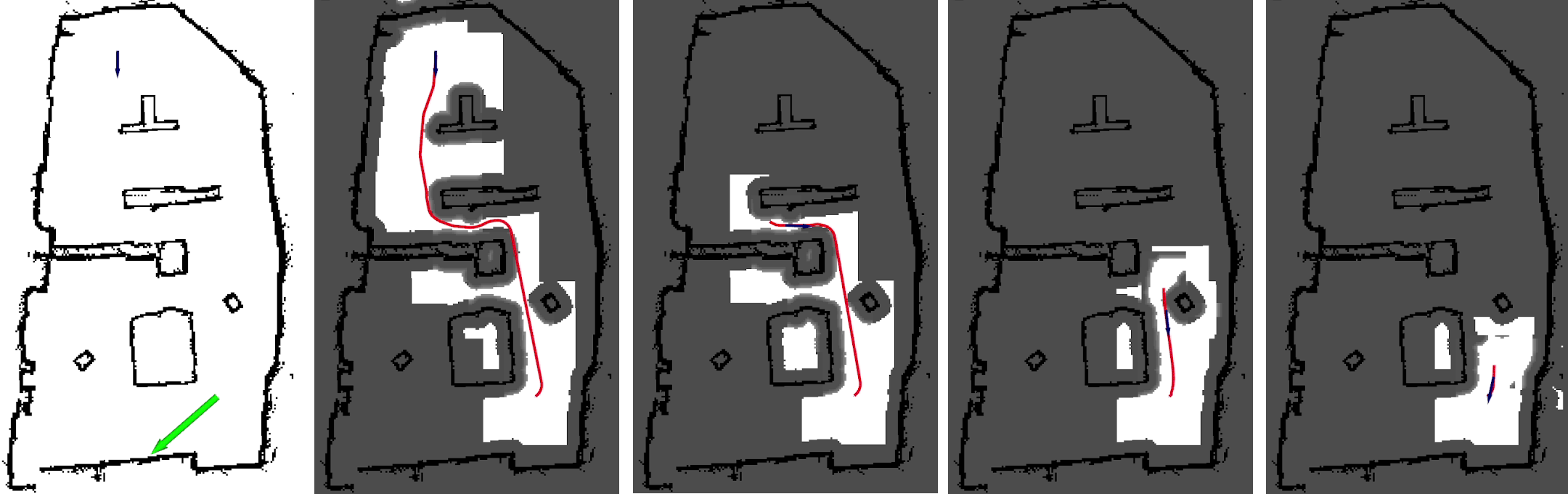}
    \vspace*{-1.5em}
    \caption{Execution of trajectory planned using MPT aided planner using the Nav2 stack. For a given goal position (green arrow), the MPT model proposes a region from the current position of the robot (blue arrow) to the goal, and the Hybrid A* planner generates a global path (red). Because of the low overhead of generating patches, the MPT model is able to aid the underlying planner in real time.}
    \label{fig:nav2PlannedPath}
    \vspace{-1em}
\end{figure*}

Due to classification errors, MPT fails on around 1\% of maps. Nevertheless, by randomly exploring the map for a few samples outside the segmented region, MPT-RRT*-EE can find valid trajectories without additional costs. UNet-RRT*-EE, on the other hand, has similar statistics to RRT*, implying that the planner relies on RRT* alone to generate a path. The distribution of planning time and vertices for MPT-RRT*-EE are also much tighter compared to traditional planners and even MPT planners without exploration.


The following experiment we conducted evaluated MPT's ability to generalize to map sizes that were not part of the training data. We tested the model on four different map sizes, 360$\times$240, 480$\times$240, 560$\times$560, and 780$\times$780 of the Random Forest environment on 1000 randomly generated maps while maintaining the density of obstacles. We used the same MPT model without any re-training or fine-tuning and compared the same metrics with traditional planning techniques. We did not compare against learning-based planners because these planners either performed poorly in our previous experiment or were not generalizable to maps of different sizes without modifications. The planning statistics are summarized in Table \ref{tab:different_mapSize}, and three successfully planned paths using MPT are shown in Fig. \ref{fig:different_mapSize}. Training the model by randomly shifting position encoding was instrumental in improving the accuracy of the planner for maps of larger sizes. Results for the model trained with fixed position encoding (F.E.) from \cite{nips_attention} are given in Table 3.  We observe that the MPT model trained with randomized position encoding achieves nearly 61\% more accuracy on larger maps while improving 1-4\% on smaller maps.



We notice that the MPT aided planners, similar to our previous experiments, achieve lower planning time and vertices count than traditional planners without additional training or fine-tuning for larger maps. IRRT* and BIT* achieves similar performance to MPT-aided planners for smaller-sized maps but as the map sizes grow, the time taken by IRRT* grows because of the more prominent search space. While for larger maps, MPT aided planners can find a solution faster and outperform IRRT* and BIT*.



\subsection{Dubins Car Model}
To evaluate the performance of MPT for the Dubins car model, we compared our method with RRT* and SST planners. For the geometric planning, we used Dubins curves \cite{Dubins_1957} as edge connectors, while for SST nodes were constructed by propagating wheel velocity and steering angle. The parameters for SST were set as reported in \cite{li2016asymptotically}. The metrics we report is the same as the point robot. The results are summarized in Table \ref{tab:dubinsCar}. We see that MPT-aided planners are able to reduce planning time and sampled points by half, while the hybrid planning strategy is able to achieve 100\% accuracy without adding additional vertices.

Fig. \ref{fig:dubins_car_map} shows two examples of paths planned using MPT for the Dubins car. We can observe that the MPT model is able to predict regions through which a kinematically feasible path can pass through. By narrowing down the search space, the planner is able to generate optimal paths with fewer samples compared to the unaided RRT* planner. A comparison for points generated between RRT* and MPT-RRT* is shown in Fig. \ref{fig:car_samp}. The SST planner on the other hand takes more time and vertices to generate a valid path because more samples are required to traverse through the narrow sections of the map.

\begin{table}
        \vspace{1em}
        \centering
        \caption{Comparing planning accuracy, and median time and vertices for Dubins Car Model for the Random Forest Environment.}
        \label{tab:dubinsCar}
        \begin{tabular}{lccc}
            \toprule
            Planner     &  Accuracy & Time (sec) & Vertices\\
            \midrule
            RRT*        & 100\%  & 0.357  & 95 \\
            SST         & 100\%  & 4.880  & 710\\
            MPT-RRT*    & 95.15\%& 0.176  & 59 \\
            MPT-RRT*-EE & 100\%  & 0.197  & 60 \\
            \bottomrule
        \end{tabular}
        \vspace{-2em}
\end{table}


\section{MPT Navigation2 Plugin}


We also provide the MPT model as a global planner plugin to the Navigation2 \cite{9341207} (Nav2) navigation stack. The model is integrated in such a way that, any of the already existing planners in Nav2 could be used to generate a plan in the accentuated area proposed by the MPT model. We tested the plugin on an RC Dubins car in an indoor mapped environment. The model used by the plugin was the one trained using just the synthetic data for the point robot. The planning frequency was set at 10Hz, and the Hybrid A* planner with Dubins motion model was used to generate the final global plan. A simple DWB local planner was used to follow the planned global path. An example of a complete execution of a planning problem is show in Fig. \ref{fig:nav2PlannedPath}.

This experiment also highlights the sim-2-real generalization of the MPT model. Previous learning models would have to gather task-specific data \cite{8412538, ichter2018learning} or actively learn from failed trajectories\cite{qureshi2020motion} in order to adapt to new environments. The ability of MPT to generalize will also benefit the entire robotics community by making it easier to use previously trained models.


\section{Conclusion} 
In the future, extending MPT to higher dimensional robots such as robotic arms or drones is an interesting problem with many real-world applications. For mobile robots in 3D space, the use of sparse Transformer models would be better suited because of the dimensionality of the space. While for manipulation systems in 3D since the task and joint space do not overlap, the extension of method is more challenging.

In this work, we have shown an application of the transformer model for 2D navigation tasks. Unlike prior methods that need to retrain models for maps of different sizes, we leverage the ability of transformers to handle sequences of different lengths and parallelize long-term dependencies without recursions for planning problems with different map sizes. We have also shown the ability of MPT to reduce search spaces for robots with kinematic constraints by predicting key points in $SE(2)$ space. Hence by combing MPT with SMP, we are able to generate paths faster, with fewer tree nodes for different environments and robot models.







\bibliographystyle{IEEEtran}
\bibliography{root}

\end{document}